\documentclass[letterpaper, 10 pt, conference]{ieeeconf}  

\IEEEoverridecommandlockouts                              

\usepackage{amsfonts}
\usepackage{amsmath} 
\usepackage{amssymb}  
\usepackage{graphicx}
\usepackage{subcaption}
\usepackage{float}
\usepackage{booktabs}
\usepackage{multirow}
\usepackage{xcolor}
\usepackage{hyperref}
\usepackage{pifont}
\newcommand{\cmark}{\ding{51}}%
\newcommand{\xmark}{\ding{55}}%

\title{\LARGE \bf

PokeNet: Learning Kinematic Models of Articulated Objects from Human Observations
}

\author{Anmol Gupta$^{1}$, Weiwei Gu$^{1}$, Omkar Patil$^{1}$, Jun Ki Lee$^{2}$, Nakul Gopalan$^{1}$
\thanks{$^{1}$School of Computation and AI, ASU, Tempe
        {\tt\small \{anmolgupta, weiweigu, opatil3, ng\}}@asu.edu}%
\thanks{$^{2}$AI Institute, Seoul National University, Seoul, South Korea
        {\tt\small junkilee@snu.ac.kr}}%
}

\newcommand\blfootnote[1]{%
  \begingroup
  \hypersetup{hidelinks}
  \renewcommand\thefootnote{}\footnote{#1}%
  \addtocounter{footnote}{-1}%
  \endgroup
}
\begin{document}

\maketitle

\thispagestyle{empty}
\pagestyle{empty}


\begin{abstract}

Articulation modeling enables robots to learn joint parameters of articulated objects for effective manipulation which can then be used downstream for skill learning or planning.
Existing approaches often rely on prior knowledge about the objects, such as the number or type of joints. Some of these approaches also fail to recover occluded joints that are only revealed during interaction. Others require large numbers of multi-view images for every object, which is impractical in real-world settings. Furthermore, prior works neglect the order of manipulations, which is essential for many multi-DoF objects where one joint must be operated before another, such as a dishwasher.
We introduce PokeNet, an end-to-end framework that estimates articulation models from a single human demonstration without prior object knowledge. Given a sequence of point cloud observations of a human manipulating an unknown object, PokeNet predicts joint parameters, infers manipulation order, and tracks joint states over time.
PokeNet outperforms existing state-of-the-art methods, improving joint axis and state estimation accuracy by an average of over $27\%$ across diverse objects, including novel and unseen categories. We demonstrate these gains in both simulation and real-world environments. 
Code and dataset are available on our webpage$^+\!\!$.

\blfootnote{$^+$Webpage: \href{https://sequential-joints.github.io/}{https://sequential-joints.github.io/}}

\end{abstract}


\section{INTRODUCTION}


\maketitle
\begin{figure*}[t]
\includegraphics[scale=0.7]{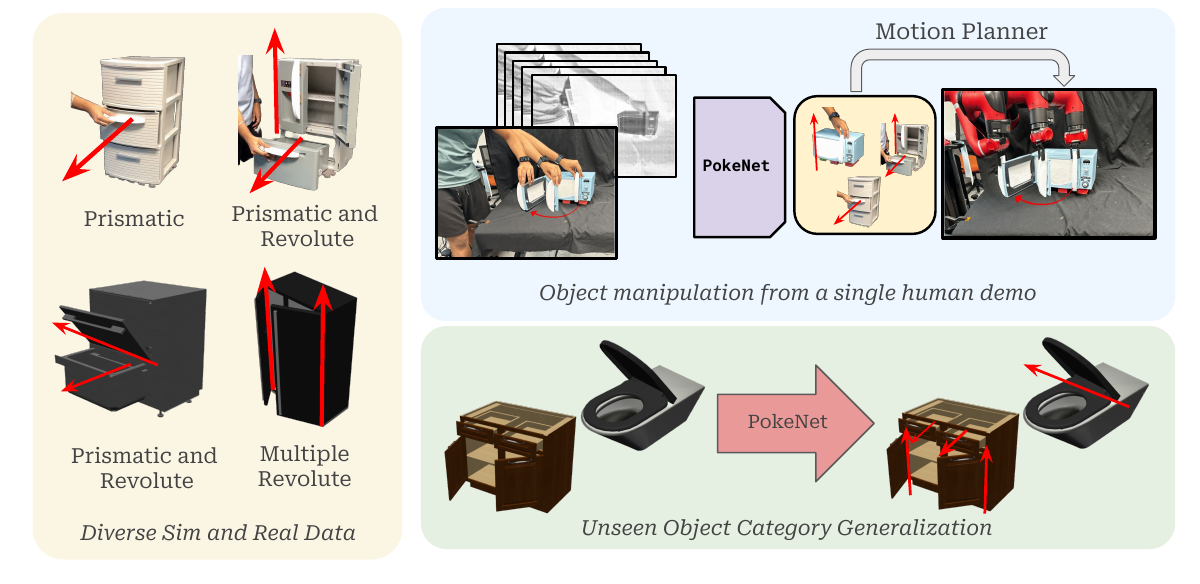}
\centering

\caption{We propose a novel framework that learns the joint parameters and manipulation order of articulated objects directly from human demonstrations. Leveraging human interactions allows our method to reason about occluded joints, while requiring no prior object knowledge. It generalizes to unseen object categories and achieves state-of-the-art performance in both simulation and real-world settings.}
\label{fig:overview}
\end{figure*}


Articulation modeling is the task of learning structured object models that capture joints, degrees of freedom (DoF), and motion constraints to provide a feasible direction for the manipulation of an articulated joint. Such models can be used by planners or policies across robot embodiments to generate feasible manipulation strategies even on unseen objects.

Many prior works on articulation modeling rely on a single RGB-D image \cite{Abbatematteo2019LearningTG}\cite{pmlr-v229-zhang23c}\cite{EisnerZhang2022FLOW}\cite{chen2024urdformer}\cite{geng2022gapartnet}. However, static views are often insufficient: joints may be occluded or revealed only through interaction, for example, a drawer hidden inside a refrigerator. Moreover, a single image provides no information about the range or limits of joint motion, which are critical for safe manipulation, for example, opening a laptop hinge without exceeding its allowable range. Other approaches, such as~\cite{Abbatematteo2019LearningTG}\cite{jain2021screwnet}\cite{jain2022distributional} achieve promising results but rely on strong assumptions, such as known number of joints or type of joints.
Recent works, inspired by 3D reconstruction methods like NeRF~\cite{mildenhall2020nerf}, NeuS\cite{wang2021neus} and Gaussian Splatting\cite{kerbl3Dgaussians}, use reconstruction to predict the articulation parameters of objects ~\cite{kim2025screwsplat}\cite{yu2025artgs3dgaussiansplattinginteractive}\cite{liu2025artgsbuildinginteractablereplicas}\cite{kerr2024rsrd}. While effective, these techniques require a large number of multi-view images and often use multistage pipelines.
Importantly, none of these methods accounts for the sequence of manipulations in multi-DoF objects, which is critical when one joint must be operated before another, such as opening a dishwasher door before pulling out its rack.

To address these limitations, we propose PokeNet, an end-to-end method for estimating articulation parameters from a single human demonstration without requiring prior knowledge of the object. By observing a person manipulate an object through a sequence of point cloud frames, PokeNet captures not only the articulation structure, including joint types and parameters of occluded joints, but also the intermediate state of each joint at every time step. In addition, it infers the order of joint operations from demonstration, which is essential for many multi-DoF objects.
A key challenge is that every novel object has an unknown number and hierarchy or ordering of joints. Traditional models assume a fixed number of outputs, making them unsuitable for this task. To tackle this problem, we adopt a set prediction formulation: the network directly outputs a set of joint hypotheses and uses a permutation invariant loss to align them with ground truth. This design avoids assumptions about joint count, while enabling end-to-end learning of articulation parameters. A qualitative comparison of PokeNet with prior works is given in Table \ref{tab:capability}.
In summary, our contributions are as follows:

\begin{itemize}
   \item In this work, we propose a novel framework for learning articulated objects with multiple degrees of freedom (DoFs) from a single human demonstration captured from one viewpoint. Our method jointly estimates the articulation structure (joint types and parameters), the state of each joint at every time step, and the correct sequence of joint manipulations required to reach a target configuration. \textbf{To our knowledge, this is the first method to simultaneously model articulation structure and manipulation order from human input for multi-DoF objects.}

\item We release a comprehensive real-world dataset of articulated objects, spanning seven objects from four categories and comprising $5{,}500$ human–object interaction data points. \textbf{To our knowledge, this is the largest annotated real-world articulated object corpus.}  

\item For evaluation, we conduct extensive experiments on simulated and real-world multi-DoF objects. Our model generalizes to four unseen categories in simulation, remains robust to variations in object sizes, and extends to unseen real-world instances. \textbf{PokeNet achieves state-of-the-art performance, improving joint axis and state estimation by up to $25\%$ on $8{,}000$ simulated data points and by $30\%$ on $1{,}600$ real-world data points.} Finally, we demonstrate successful manipulation on both seen and unseen object classes using a motion planner.
   
\end{itemize}

    

\begin{table}
\centering
\caption{Capability comparison of articulation modeling methods. PokeNet is the only method that jointly handles multi-joint objects, recovers occluded or shut joints, and predicts both joint states and manipulation order.}
\begin{tabular}{lccccc}
\hline
Method & Sng. & Multi & Occ. & State & Order \\
\hline
ScrewNet  & \cmark & \xmark & \xmark & \cmark & \xmark \\
GAPartNet & \cmark & \cmark & \xmark & \xmark & \xmark \\
\textbf{PokeNet}   & \cmark & \cmark & \cmark & \cmark & \cmark \\
\hline
\end{tabular}
\label{tab:capability}
\end{table}

\section{Related Work}

\textbf{Estimation from Visual Data.} Several approaches estimate the articulation properties of objects directly from visual input~\cite{Abbatematteo2019LearningTG,jain2021screwnet,jain2022distributional}. However, these methods often assume prior object category knowledge or are limited to single-DoF objects. Other approaches~\cite{li2020categorylevelarticulatedobjectpose,10.1145/3272127.3275027} rely on predefined models or DoF assumptions, restricting generalization. More recent methods~\cite{pmlr-v229-zhang23c,EisnerZhang2022FLOW,geng2022gapartnet} take a single partial point cloud as input, but such approaches cannot reliably recover occluded or fully closed joints and provide no information on joint motion limits, posing safety risks when deployed on real robots. In contrast, our method makes no assumptions about object categories or joint types, requiring only that the links be rigid and that a maximum number of joints be specified. Using full motion sequences from a single human demonstration, PokeNet is able to recover occluded joints, capture motion limits, and enable safer deployment in real-world settings.


\textbf{Interactive Methods.} Interactive perception approaches~\cite{4543220, Katz-2013-7694, 7487714, nie2022sfa} involve robot-driven exploration to estimate articulation. However, these methods require object textures and rely on primitive actions like pushes or pokes, limiting their effectiveness in complex settings. We use human demonstrations to overcome these limitations, capturing more realistic interaction dynamics without requiring object texture or category priors. Our method generalizes across diverse object instances and handles occluded configurations more robustly.

\textbf{Reconstruction-Based Methods.} Several approaches recover articulation models through reconstruction frameworks such as NeRF and Gaussian Splatting~\cite{jiayi2023paris,weng2024neural,liu2025artgsbuildinginteractablereplicas}. These methods often rely on strong priors about object categories or structures, limiting their generalization. More recent work such as ScrewSplat~\cite{kim2025screwsplat} removes some of these assumptions but still requires large numbers of multi-view images to estimate articulation parameters. In contrast, our method requires no prior object knowledge and predicts articulation parameters from human demonstration. PokeNet is designed to infer articulation parameters with sample efficiency, without reconstructing the full visual representation of the object.

\section{Problem Formulation}

Articulated objects have components that are connected at joints. These components are called ``links'' that can be rigid. The ``joints'' afford relative motion between these links. The joints of these objects can be classified into two broad classes – revolute and prismatic joints. These joints can be defined using a direction and an anchor point in space. For prismatic joints, the link moves along the direction of the axis, while for revolute joints, the link moves in a direction perpendicular to the axis. The axis and position of the joint together are called articulation parameters.  

In our problem formulation, we assume the links are rigid and the maximum number of joints are $K$. The input is a sequence of 3D point clouds of the object while it is being manipulated. Formally, each data point is a sequence $\mathcal{P} = \{ P_1, P_2, \ldots, P_T \}, \quad P_t \in \mathbb{R}^{N \times 3},$
where $T$ is the number of observed frames and $N$ is the number of 3D points per frame.

The objective is to recover the underlying articulation model directly from \(\mathcal{P}\). 
We define a prediction function parameterized by \(\theta\):
\[
f_\theta : \mathcal{P} \mapsto \mathcal{S} = \{ s_1, s_2, \ldots, s_K \},
\]
where \(\mathcal{S}\) is a set of \(K\) slots. Each slot corresponds to a hypothesized joint and is defined as
\[
s_k = \{ c_k, \tau_k, \mathbf{d}_k, \mathbf{p}_k, o_k \}.
\]
Here, \(c_k \in [0,1]\) is a confidence score that determines if slot \(s_k\) corresponds to a valid joint; \(\tau_k \in \{0,1\}\) denotes the joint type (\(0\) revolute, \(1\) prismatic); \(\mathbf{d}_k \in \mathbb{R}^3\) is a unit axis direction; \(\mathbf{p}_k \in \mathbb{R}^3\) is an anchor point on this axis; and \(o_k \in [0,1]\) encodes the relative manipulation order.  

In addition to the articulation parameters, we predict per-timestep joint states aligned with the same slots. Let
$
\mathcal{Y} = \{ \mathbf{y}_1, \ldots, \mathbf{y}_T \}, \quad \mathbf{y}_t \in \mathbb{R}^{K \times3},
$
denote the auxiliary output, where the \(k\)-th component \(y_{t,k}\) represents the displacement of slot \(s_k\) at time \(t\). Its physical interpretation depends on the type: for $\tau_k=0$ (revolute), $y_{t,k} = (\sin\theta_{t,k},\cos\theta_{t,k},0)$ encodes angular displacement in a wraparound-free form, while for $\tau_k=1$ (prismatic), $y_{t,k} = (0,0,\rho_{t,k})$ encodes linear displacement.

The final output is obtained by keeping only slots with high confidence,
\[
\hat{\mathcal{S}} = \{ s_k \in \mathcal{S} \;|\; c_k > \mu \},
\]
and taking their associated per-timestep states \(\hat{\mathcal{Y}} = \{ y_{t,k} \mid s_k \in \hat{\mathcal{S}},\; t=1\ldots T \}\). The retained set is sorted by \(o_k\) to provide an ordered articulation model together with its time-varying joint states.

\section{Dataset}

For the proposed model, we needed an extensive range of objects to learn from. For this, we used two distinct datasets: a simulated dataset and a real-world dataset. 

\subsection{Simulated Dataset}   
The simulated dataset was created using the Partnet-Mobility dataset ~\cite{Xiang_2020_SAPIEN, chang2015shapenet, Mo_2019_CVPR}, which we chose for its diversity and ease of use, allowing us to test the robustness and generalizability of our approach under controlled conditions.
Each object was rendered in simulation and its articulated links were moved to simulate realistic joint motions. We captured this motion as sequences of point clouds in the camera coordinate frame, which were then used to predict articulation parameters.

In total, we collected $110{,}000$ sequences spanning eleven categories: microwave, laptop, washing machine, fridge, drawer, bucket, door, fan, scissors, window, and trashcan. These objects were chosen for their variability in size, geometry, and degrees of freedom, as well as their practical relevance in real-world settings. We additionally reserved four categories—furniture, box, toilet, and pliers—exclusively for testing. This set of objects is comparable to those used in prior baselines.

\subsection{Real World Object Dataset with Ground Truth}
Due to the absence of large-scale real-world 3D video datasets for articulated objects, we collected our own. We selected four commonly encountered household objects—microwave, dishwasher, refrigerator, and drawer—chosen for their availability and diversity in joint structure. These categories collectively cover a wide range of articulation types, including both revolute and prismatic joints. For example, microwaves exhibit standard revolute joints, while dishwashers and refrigerators often feature combinations of revolute and prismatic links.

Each object was physically manipulated by three of the authors by opening and closing its movable parts, and the resulting joint displacements and angles were recorded. To obtain accurate ground-truth articulation data, we attached ArUco markers \cite{garrido2014automatic} to object parts and tracked their motion with a dual calibrated camera setup, ensuring each marker was visible from at least one camera throughout the sequence. Note that this two-camera setup was used only for annotation; the demonstration videos used for training were recorded from a single camera view.

In total, we collected $5{,}500$ annotated point cloud sequences across four object categories. Of these, $3{,}900$ samples were used for training and $1{,}600$ for testing, making this, to the best of our knowledge, the largest real-world dataset of 3D videos of articulated objects annotated with ground-truth joint parameters. To assess generalization, we also recorded $10$ physical demonstrations each for two additional unseen objects: a slider knife and a stapler.

\section{Methods}

In this work, we propose a novel end-to-end framework for estimating the articulation parameters and manipulation order of articulated objects with multiple degrees of freedom (DoFs) using sequential point cloud observations from a \textit{single camera view}.
For a novel object, the robot cannot know in advance either the number of joints or their manipulation order. This requires predictions to be made over \emph{sets} of joints with no predefined order. To address this, we adopt a DETR-style decoder~\cite{10.1007/978-3-030-58452-8_13} with Hungarian matching, which aligns predicted slots with ground-truth joints in a permutation-invariant manner.

Our method begins by encoding the input sequence of point clouds with a PointNet++ encoder~\cite{qi2017pointnetplusplus} and a lightweight transformer encoder~\cite{vaswani2017attention}, which produce spatial embeddings for each frame. The sequence of embeddings is then processed by another transformer encoder to capture temporal dependencies across frames. The resulting temporal features are passed into a transformer decoder that maintains a fixed set of learnable queries, each of which attends to the sequence and specializes into a “slot” representing a potential joint. 

Each slot outputs: a \emph{confidence score} indicating whether it corresponds to a valid joint, a \emph{joint type} (revolute or prismatic), a normalized 3D \emph{axis direction}, a 3D \emph{anchor point} defining the axis position in space, and an \emph{order score} representing the relative sequence in which joints are manipulated. During inference, slots with high confidence scores are kept and sorted by their predicted order scores, producing an ordered set of joints that parametrizes the articulation model of the object.

This design provides several advantages: (i) it removes the need for prior knowledge of object categories or exact joint counts, requiring only a specified maximum number of joints, (ii) it uses temporal reasoning by attending to embeddings across frames, and (iii) it uses a structured slot-based representation that is generalizable across objects with varying articulations.

\subsection{Architecture}

In this section, we describe the architecture of \textbf{PokeNet}, which consists of three main components: (i) spatial encodings, (ii) temporal encodings, and (iii) decoders for joint parameter prediction.

\textbf{Spatial Encodings.}
Each frame $P_t$ is a 3D point cloud. We encode it with PointNet++~\cite{qi2017pointnetplusplus} to obtain $Q$ point-level embeddings
$X_t=\{x_{t,1},\ldots,x_{t,Q}\}\in\mathbb{R}^{Q\times D}$.
We append a learnable $[\texttt{CLS}]$ token $c_t$ and pass $\{c_t\}\cup X_t$ through a lightweight transformer encoder~\cite{vaswani2017attention}.
The updated $[\texttt{CLS}]$ output $\hat c_t\in\mathbb{R}^{D}$ serves as the \emph{frame-level representation}, summarizing the frame’s information via attention over the local point tokens.

\textbf{Temporal Encodings.}  
The sequence $\{\hat{c_0}, \hat{c_1}, \ldots, \hat{c_{T-1}}\}$ is then processed by a transformer encoder~\cite{vaswani2017attention} to capture temporal dependencies. 
These temporal features form the memory that is accessed by the decoders in the next stage.

\textbf{Joint Set Decoder.}  
The first decoder is responsible for predicting the articulation parameters. We adopt a DETR-style formulation \cite{10.1007/978-3-030-58452-8_13}, where a fixed set of $K$ learnable queries attends to the temporal memory. Each query is updated through cross attention and specializes in a ``slot'' corresponding to a potential joint. For every slot, we attach prediction heads that output: a confidence score, a joint type (revolute or prismatic), a normalized axis direction $\mathbf{d} \in \mathbb{R}^3$, an anchor point $\mathbf{p} \in \mathbb{R}^3$, and an order score $o \in [0,1]$ indicating the relative sequence in which the joint is manipulated. The slot-based formulation allows the model to predict a variable number of joints under $K$.

\textbf{Auxiliary Decoder.} In addition to the joint set decoder, PokeNet includes an auxiliary decoder that tracks the per-frame state of each joint across the observed sequence. For every slot $s_k$ and time step $t$, the decoder outputs a state vector $\mathbf{z}_{t,k} \in \mathbb{R}^3$. The three components are interpreted according to the slot’s joint type: for revolute joints, $\mathbf{z}_{t,k} = (\sin\theta_{t,k}, \cos\theta_{t,k}, 0)$ encodes angular displacement in a wraparound-free form, while for prismatic joints, $\mathbf{z}_{t,k} = (0, 0, \rho_{t,k})$ provides linear translation. This unified representation avoids discontinuities for angles while maintaining a consistent $K \times T \times 3$ output structure across all slots. In the final output, only states associated with slots that exceed the confidence threshold $c_k > \mu$ are retained, ensuring that per-frame states are defined solely for valid joints.

\begin{figure*}[t]
\includegraphics[width=0.98\textwidth]{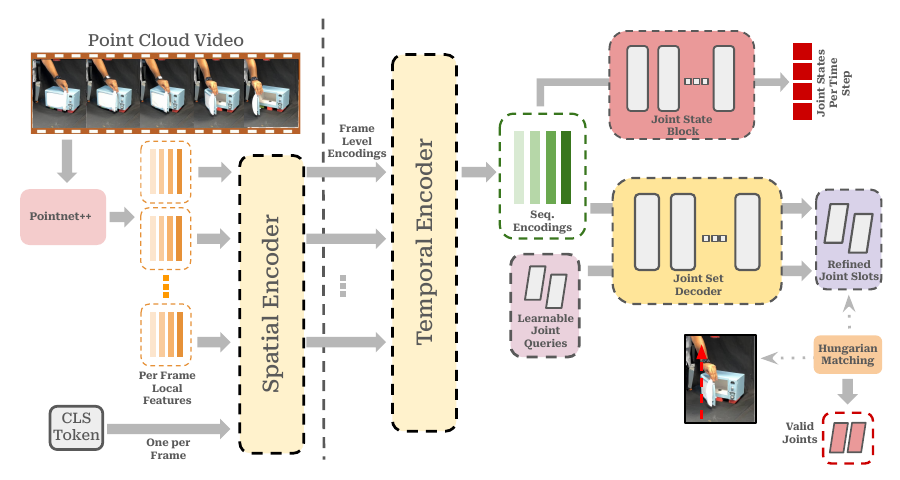}
\centering
\caption{Overview of \textbf{PokeNet}. Our model takes a sequence of point clouds as input. Each frame is encoded with PointNet++ to extract spatial features, and [CLS] tokens are passed through a transformer encoder to capture temporal dependencies. A DETR style joint decoder with learnable queries attends to these temporal features to predict an ordered kinematic slot model, including joint type, axis direction, anchor point, confidence, and manipulation order. An auxiliary decoder additionally predicts per-frame joint states.}
\label{fig:architecture}
\end{figure*}

\begin{table*}[h!]
\renewcommand{\arraystretch}{1.2}
\centering
\begin{tabular}{l|l|ccc}
\toprule
Object & Method & Axis Orientation Error (°) & Axis Displacement Error (cm) & Joint State Error (deg/cm) \\
\midrule
Microwave
  & Screwnet     & 21.13 $\pm$ 1.07 & 14.28 $\pm$ 1.03 & 18.47 $\pm$ 0.96 \\
  & GAPartNet    & 9.71 $\pm$ 1.12 &  6.37 $\pm$ 1.21 & \textbf{-} \\
  & Ours& \textbf{7.26 $\pm$ 1.08} & 5.26 $\pm$ 0.97 & \textbf{9.31 $\pm$ 1.09} \\
\midrule
Washing Machine
  & Screwnet     & 21.78 $\pm$ 0.88 & 11.64 $\pm$ 1.22 & 21.36 $\pm$ 1.27 \\
  & GAPartNet    & 10.57 $\pm$ 1.33 &  7.92 $\pm$ 1.08 & \textbf{-} \\
  & Ours& \textbf{7.48 $\pm$ 1.01} & \textbf{6.23 $\pm$ 1.15} & \textbf{7.11 $\pm$ 1.34} \\
\midrule
Laptop
  & Screwnet     & 22.69 $\pm$ 1.14 & 12.37 $\pm$ 0.91 & 17.42 $\pm$ 1.38 \\
  & GAPartNet    & 7.28 $\pm$ 0.95 &  5.72 $\pm$ 1.44 & \textbf{-} \\
  & Ours& 6.14 $\pm$ 1.23 & 4.68 $\pm$ 1.07 & \textbf{8.21 $\pm$ 1.07} \\
\midrule
Fridge
  & Screwnet     & 26.87 $\pm$ 1.25 & 13.58 $\pm$ 1.09 & 24.21 $\pm$ 1.11 \\
  & GAPartNet    &  11.74 $\pm$ 1.18 &  8.83 $\pm$ 0.86 & \textbf{-} \\
  & Ours& \textbf{8.92 $\pm$ 0.97} & \textbf{6.77 $\pm$ 1.31} & \textbf{9.48 $\pm$ 0.91} \\
\midrule
Drawer
  & Screwnet     & 23.58 $\pm$ 0.93 & 10.96 $\pm$ 1.36 & 20.66 $\pm$ 0.84 \\
  & GAPartNet    & 10.07 $\pm$ 1.09 &  6.77 $\pm$ 1.12 &  \textbf{-} \\
  & Ours& \textbf{7.36 $\pm$ 1.14} & \textbf{4.59 $\pm$ 0.88} & \textbf{6.91 $\pm$ 1.12} \\
\midrule
Trashcan
  & Screwnet     & 24.72 $\pm$ 1.31 & 14.61 $\pm$ 0.99 & 19.18 $\pm$ 1.43 \\
  & GAPartNet    & 9.34 $\pm$ 0.86 &  7.37 $\pm$ 1.27 & \textbf{-} \\
  & Ours& 8.51 $\pm$ 1.08 & \textbf{4.92 $\pm$ 1.13} & \textbf{10.03 $\pm$ 1.36} \\
\midrule
Window
  & Screwnet     & 20.93 $\pm$ 1.17 & 12.83 $\pm$ 1.02 & 21.57 $\pm$ 0.98 \\
  & GAPartNet    & 9.39 $\pm$ 1.24 &  6.58 $\pm$ 1.14 & \textbf{-} \\
  & Ours& \textbf{7.16 $\pm$ 0.89} & 5.73 $\pm$ 1.41 & \textbf{6.88 $\pm$ 1.05} \\
\midrule
Door
  & Screwnet     & 24.05 $\pm$ 1.08 & 13.47 $\pm$ 1.20 & 23.62 $\pm$ 1.26 \\
  & GAPartNet    & 9.88 $\pm$ 1.43 &  8.11 $\pm$ 0.92 & \textbf{-} \\
  & Ours& 9.03 $\pm$ 1.12 & \textbf{5.06 $\pm$ 1.33} & \textbf{7.95 $\pm$ 1.14} \\
\midrule
Fan
  & Screwnet     & 22.41 $\pm$ 0.99 & 11.72 $\pm$ 1.28 & 20.84 $\pm$ 1.21 \\
  & GAPartNet    &  9.93 $\pm$ 1.02 &  7.68 $\pm$ 1.34 & \textbf{-} \\
  & Ours& \textbf{7.84 $\pm$ 0.95} & \textbf{4.51 $\pm$ 1.07} & \textbf{6.59 $\pm$ 1.18} \\
\midrule
Scissors
  & Screwnet     & 21.62 $\pm$ 1.34 & 10.73 $\pm$ 0.90 & 24.39 $\pm$ 1.32 \\
  & GAPartNet    & 8.85 $\pm$ 1.20 &  8.21 $\pm$ 1.25 & \textbf{-} \\
  & Ours& \textbf{6.12 $\pm$ 1.06} & \textbf{5.87 $\pm$ 0.82} & \textbf{8.36 $\pm$ 1.06} \\
\midrule
Bucket
  & Screwnet     & 20.74 $\pm$ 0.92 & 12.42 $\pm$ 1.35 & 22.53 $\pm$ 1.09 \\
  & GAPartNet    & 10.66 $\pm$ 1.26 &  6.97 $\pm$ 1.10 &  \textbf{-} \\
  & Ours& \textbf{8.21 $\pm$ 0.96} & \textbf{4.38 $\pm$ 1.22} & \textbf{7.44 $\pm$ 1.03} \\
\midrule
Plier* 
  & Screwnet     & 27.84 $\pm$ 1.19 & 13.66 $\pm$ 0.88 & 19.47 $\pm$ 1.31 \\
  & GAPartNet    & 10.63 $\pm$ 1.08 &  6.94 $\pm$ 1.12 & \textbf{-}\\
  & Ours& \textbf{8.14 $\pm$ 1.23} & 6.58 $\pm$ 1.29 & \textbf{11.37 $\pm$ 0.99} \\
\midrule
Toilet* 
  & Screwnet     & 29.31 $\pm$ 1.15 & 14.12 $\pm$ 1.33 & 21.66 $\pm$ 0.97 \\
  & GAPartNet    & 12.08 $\pm$ 0.96 &  7.76 $\pm$ 1.18 & \textbf{-} \\
  & Ours& \textbf{8.52 $\pm$ 1.01} & \textbf{5.23 $\pm$ 0.93} & \textbf{9.97 $\pm$ 1.09} \\
\midrule
Furniture* 
  & Screwnet     & 26.47 $\pm$ 1.28 & 11.95 $\pm$ 1.07 & 25.88 $\pm$ 1.44 \\
  & GAPartNet    & 17.24 $\pm$ 1.11 &  8.62 $\pm$ 0.91 & \textbf{-}\\
  & Ours& \textbf{13.15 $\pm$ 0.90} & \textbf{5.87 $\pm$ 1.36} & \textbf{10.94 $\pm$ 1.27} \\
\midrule
Box* 
  & Screwnet     & 25.72 $\pm$ 0.91 & 12.31 $\pm$ 1.24 & 26.18 $\pm$ 0.84 \\
  & GAPartNet    & 19.37 $\pm$ 1.36 &  7.23 $\pm$ 1.06 & \textbf{-} \\
  & Ours& \textbf{11.42 $\pm$ 1.07} & \textbf{5.11 $\pm$ 0.95} & \textbf{13.82 $\pm$ 1.38} \\
\bottomrule
\end{tabular}
\caption{Results for simulated dataset. We report the mean error (with 95\% confidence intervals) for joint axis directions (in degrees), positions (in centimeters), and joint state estimation (in deg/cm), averaged over the available axes per object. All the objects were in partially opened state for GAPartNet. * represents objects unseen during training.}
\label{tab:sim-table-avg}
\end{table*}

\begin{table*}[t]
\renewcommand{\arraystretch}{1.2}
\centering
\begin{tabular}{l|l|ccc}
\toprule
Object & Method & Axis Orientation Error (°) & Axis Displacement Error (cm) & Joint State Error (deg/cm)\\
\midrule
Microwave
  & Screwnet     & 28.42 $\pm$ 1.36 & 17.58 $\pm$ 1.22 & 29.11 $\pm$ 1.09 \\
  & GAPartNet    & 18.62 $\pm$ 1.12 & 12.77 $\pm$ 1.31 & \textbf{-} \\
  & Ours& \textbf{13.88 $\pm$ 0.96} & 11.42 $\pm$ 1.24 & \textbf{13.15 $\pm$ 1.12} \\
\midrule
Fridge
  & Screwnet     & 30.97 $\pm$ 1.07 & 31.66 $\pm$ 1.38 & 27.54 $\pm$ 1.41 \\
  & GAPartNet    & 21.41 $\pm$ 1.46 & 20.82 $\pm$ 0.98 & \textbf{-}\\
  & Ours& \textbf{16.97 $\pm$ 1.05} & \textbf{16.34 $\pm$ 1.27} & \textbf{12.59 $\pm$ 1.16} \\
\midrule
Drawer
  & Screwnet     & 35.33 $\pm$ 1.28 & 33.85 $\pm$ 1.11 & 25.62 $\pm$ 0.94 \\
  & GAPartNet    & 22.88 $\pm$ 0.91 & 18.41 $\pm$ 1.32 & \textbf{-} \\
  & Ours& \textbf{17.62 $\pm$ 1.39} & \textbf{13.04 $\pm$ 1.18} & \textbf{12.71 $\pm$ 1.08} \\
\midrule
Dishwasher
  & Screwnet     & 30.84 $\pm$ 0.99 & 28.87 $\pm$ 1.42 & 32.18 $\pm$ 1.22 \\
  & GAPartNet    & 23.73 $\pm$ 1.18 & 26.28 $\pm$ 1.45 & \textbf{-} \\
  & Ours& \textbf{20.44 $\pm$ 1.12} & \textbf{21.72 $\pm$ 1.05} & \textbf{20.06 $\pm$ 1.30} \\
\midrule
Slider Knife*
  & Screwnet     & 28.22 $\pm$ 3.43 & 17.47 $\pm$ 3.18 & \textbf{-} \\
  & GAPartNet    & 58.64 $\pm$ 4.09 & 9.42 $\pm$3.82 & \textbf{-} \\
  & Ours& \textbf{14.18 $\pm$ 3.25} & \textbf{6.38 $\pm$ 4.14} & \textbf{-} \\
\midrule
Stapler*
  & Screwnet     & 36.81 $\pm$ 4.32 & 16.34 $\pm$ 3.41 & \textbf{-} \\
  & GAPartNet    & 19.17 $\pm$ 5.15 & 13.04 $\pm$ 3.92 & \textbf{-} \\
  & Ours& 16.26 $\pm$ 4.98 & \textbf{8.19 $\pm$ 3.36} & \textbf{-} \\
\bottomrule
\end{tabular}
\caption{Results for real world dataset. We report the mean error (with 95\% confidence intervals) for joint axis directions (in degrees), positions (in centimeters), and joint state estimation (in deg/cm), averaged over the available axes per object. All the objects were in partially opened state for GAPartNet. * represents objects unseen during training.}
\label{tab:real-table-avg}
\end{table*}

\subsection{Loss Function}

We train PokeNet with a composite objective that supervises all aspects of the predicted joint slots. Since our formulation treats joint estimation as a set prediction task, we first align predicted slots with ground-truth joints using the Hungarian algorithm to find the assignment that minimizes a cost composed of axis direction, anchor consistency, and joint type classification. The matched pairs are then used to compute the loss.  

For each predicted slot, we supervise five components. First, an \emph{confidence loss} is applied using binary cross-entropy, with matched slots labeled as positive and unmatched slots as negatives. Second, joint \emph{type classification} (revolute vs. prismatic) is trained with cross-entropy loss. Third, the predicted axis direction is supervised using an angular loss  
$L_{\text{axis}} = \mathbb{E}[1 - |\langle \hat{\mathbf{d}}, \mathbf{d} \rangle|],$
which encourages alignment between normalized predicted and ground-truth directions while being invariant to axis sign. Fourth, anchor point prediction is trained with mean squared error, minimizing the point-to-line distance between predicted anchors and ground-truth joint axes. Fifth, we supervise the \emph{order score} predicted by each slot. 
Since ground-truth joint order is given as a discrete index 
$r \in \{0,1,\dots,M-1\}$ for $M$ joints, we map it to a continuous 
target in $[0,1]$ to enable gradient-based regression. Specifically, 
we define the normalized order as
\begin{equation}
\tilde{r} = \frac{r}{\max(1,\,M-1)} .
\end{equation}
This ensures that the first joint always maps to $0$, the last joint to $1$, 
and intermediate joints are evenly spaced between them, regardless of $M$. 
The predicted order score $\hat{o}_i \in [0,1]$ for joint $i$ is then trained 
with two complementary losses. First, an $L_1$ regression loss anchors the 
prediction to its normalized target:
\begin{equation}
L_{\text{L1}} = \frac{1}{M} \sum_{i=1}^{M} \big|\hat{o}_i - \tilde{r}_i\big| .
\end{equation}
Second, a pairwise ranking hinge loss enforces relative ordering: for any 
two ground-truth joints $i \prec j$ we require 
$\hat{o}_i + m \leq \hat{o}_j$ with margin $m$:
\begin{equation}
L_{\text{rank}} = \frac{1}{|\mathcal{X}|} 
\sum_{(i,j)\in \mathcal{X}} 
\max\big(0,\, m - (\hat{o}_j - \hat{o}_i)\big),
\end{equation}
where $\mathcal{X}$ denotes the set of all ordered joint pairs.

Finally, we introduce a state loss $L_{\text{state}}$ that applies mean squared error over the predicted per-timestep joint displacements. $L_{\text{state}}$ is computed on matched slots only, with supervision applied according to the ground-truth joint type. For revolute joints, the auxiliary decoder outputs $(\hat{u},\hat{v})$, normalized to unit length and compared against the ground-truth $(\sin\theta,\cos\theta)$. For prismatic joints, the third channel encodes the scalar displacement $\rho$, which is directly regressed. The unused channels are masked in each case.

The total loss is a weighted combination:
\begin{multline}
\mathcal{L} = \lambda_{\text{conf}} L_{\text{conf}} 
+ \lambda_{\text{type}} L_{\text{type}} 
+ \lambda_{\text{axis}} L_{\text{axis}} \\
+ \lambda_{\text{point}} L_{\text{point}} 
+ \lambda_{\text{order}} L_{\text{L1}} 
+ \lambda_{\text{rank}} L_{\text{rank}} 
+ \lambda_{\text{state}} L_{\text{state}}.
\end{multline}



\section{Experiments}

The following sections present a comprehensive evaluation of PokeNet on simulated and real-world datasets. We assess performance across four key aspects: (1) comparison with existing SOTA baselines of GAPartNet~\cite{geng2022gapartnet} and an extended multi-joint version of ScrewNet~\cite{jain2021screwnet} for joint parameters; (2) generalization to \textit{unseen object instances}; (3) generalization to \textit{unseen object categories} in both simulation and the real world; and (4) robustness to variations in \textit{object scales}, including test time scaling not seen during training.



\begin{figure*}[]
    \centering
    \begin{subfigure}{0.24\textwidth}
        \centering
        \includegraphics[width=\columnwidth, page=1]{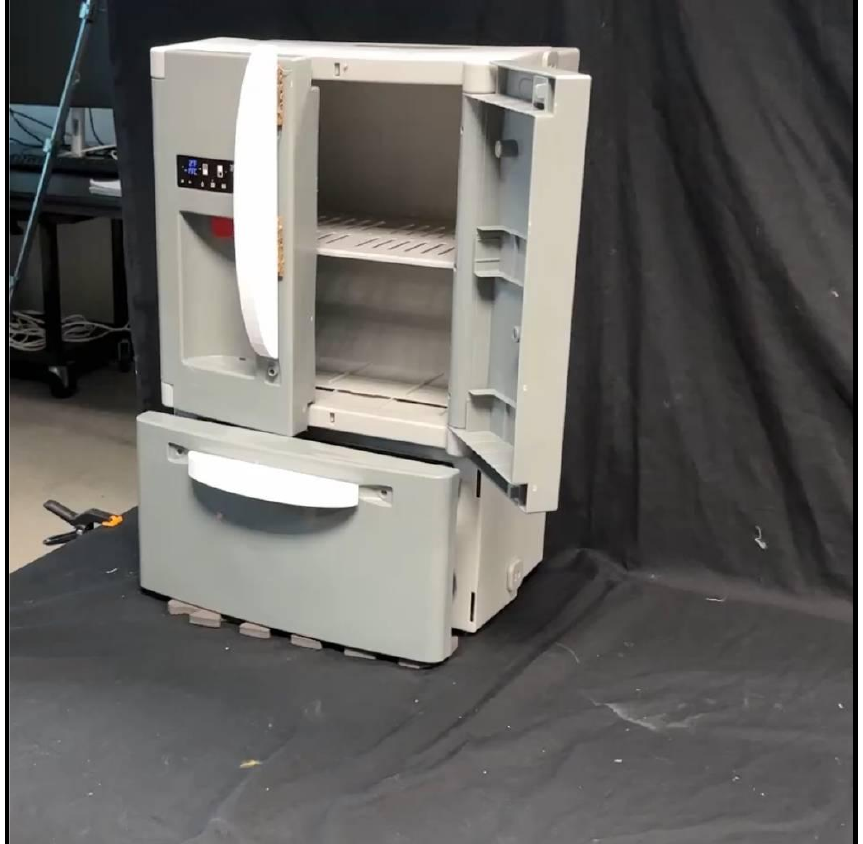}
        \caption{}
        \label{fig:sub1}
    \end{subfigure}
    \begin{subfigure}{0.24\textwidth}
        \centering
        \includegraphics[width=\columnwidth, page=2]{figures/demo/demo.pdf}
        \caption{}
        \label{fig:sub2}
    \end{subfigure}
    \begin{subfigure}{0.24\textwidth}
        \centering
        \includegraphics[width=\columnwidth, page=3]{figures/demo/demo.pdf}
        \caption{}
        \label{fig:sub3}
    \end{subfigure}
    \begin{subfigure}{0.24\textwidth}
        \centering
        \includegraphics[width=\columnwidth, page=4]{figures/demo/demo.pdf}
        \caption{}
        \label{fig:sub4}
    \end{subfigure}
    \caption{This figure shows Sawyer robot manipulating the two joints of the fridge in the order of demonstration estimated by Pokenet. (a) and (b) shows human demonstrations while (c) and (d) shows robot manipulating the object.}
    \label{fig:demo}
\end{figure*}

\subsection{Simulated Dataset Results}
From the simulated data, $88,000$
sequences were used for training, and $2,000$ per category
were set aside for testing. To evaluate category level
generalization, we additionally held out four categories
from training: furniture, box, toilet, and plier, with 2,000
sequences per category for testing.
To evaluate robustness under distribution shift, we varied
object scale during training and testing. During training,
the scales of the objects were randomly perturbed by
$\pm{5\%}$. At test time, we used unseen scales of up to $\pm{7\%}$, 
allowing us to assess robustness to scale variations beyond
the training range.

Across $30{,}000$ test samples, PokeNet significantly outperforms baseline methods in both axis alignment and anchor point accuracy. As shown in Table~\ref{tab:sim-table-avg}, it generalizes well to unseen object instances, categories, and appearance variations. On simulated data, PokeNet surpasses GAPartNet by $25\%$ in joint axis prediction accuracy.

\subsection{Real-World Dataset Results}

Across $1{,600}$ test samples, PokeNet consistently outperforms both baselines across all the object categories. PokeNet improves the accuracy of the prediction of the joint axis, exceeding GAPartNet by $30\%$. We further evaluate PokeNet on two unseen objects, a slider knife and a stapler, using 10 test demonstrations, and calculate the mean error in predicted axes. In these objects, PokeNet achieves an improvement of $40\%$ compared to GAPartNet. The results for real world data are shown in Table \ref{tab:real-table-avg}

To further validate the applicability of our method, we integrated PokeNet’s output into a planning pipeline. The robot uses the predicted articulation parameters and provided contact points to generate manipulation trajectories. For prismatic joints, the planner traces linear paths in $1$ cm increments, while for revolute joints, it traces arcs in $1^\circ$ increments. Fig.~\ref{fig:demo} shows an example of the robot successfully opening a refrigerator using parameters inferred by PokeNet. The accompanying video and website further demonstrate the robot manipulating common kitchen appliances in real world using our model’s predictions.

\begin{figure}[t]
    \centering
    \includegraphics[width=\columnwidth]{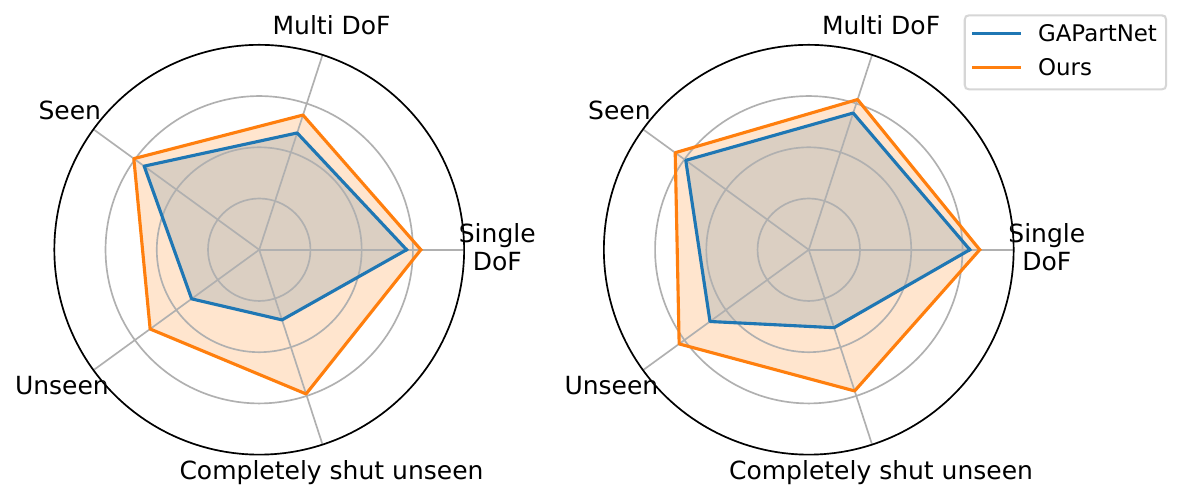}
    \caption{Comparison of PokeNet and GAPartNet on different object categories. Left: axis direction accuracy; Right: axis displacement accuracy (higher is better). GAPartNet struggles with fully shut parts, while PokeNet uses human demonstrations to generalize robustly.}
    \label{fig:radar}
\end{figure}

\subsection{Discussion}
Fig.~\ref{fig:radar} presents a radar chart comparison between GAPartNet and PokeNet across object conditions. As shown in Table~\ref{tab:capability}, PokeNet is the only method capable of jointly predicting both single and multi DoF objects, recovering occluded or fully shut joints, and estimating joint states and manipulation order by using human demonstrations. To the best of our knowledge, PokeNet is the first unified framework that integrates all these aspects of articulated object understanding.

In contrast, GAPartNet performs poorly in unseen categories with fully closed links. Its reliance on a single partial point cloud often causes it to miss parts when joints are occluded or objects are completely shut. This issue is especially worse in high-DoF objects such as storage furniture and boxes, or in objects with unusual geometries like slider knives, leading to missed segmentations and incorrect articulation estimates. PokeNet avoids these failures by using human demonstrations, which naturally reveal occluded joints and allow the model to capture articulation structure more reliably.


\section{Limitations and Future Work}

Although our method demonstrates state-of-the-art performance in estimating joint parameters, it has several limitations. First, it does not estimate contact points for manipulation. Moreover, the model does not consider obstacles, relying solely on inverse kinematics for robot motion, which increases the risk of collisions. Finally, it does not recover full object geometry, which can be useful for generating digital twins. Future work will address these gaps by detecting contact points and integrating collision-aware planning.

\section{Conclusion}
In this work, we present a novel framework that learns the kinematic constraints and manipulation sequences of multi-DoF objects directly from human demonstrations. Our approach surpasses state-of-the-art articulated object modeling methods on simulated dataset by $25\%$ and on real-world dataset by $30\%$. In addition, we contribute a new annotated real-world dataset of articulated objects that captures human interactions. Unlike prior methods, our framework makes no restrictive assumptions about the number of degrees of freedom or object categories, and requires no prior object knowledge beyond a specified maximum limit. Finally, we demonstrate that robots can leverage the learned representations to successfully manipulate diverse novel objects in real-world settings.

\section{Acknowledgment}
This material is based upon work supported by the Air Force Office of Scientific Research under award number FA9550-24-1-0239.










\bibliographystyle{IEEEtran}
\bibliography{ref}

@inproceedings{jain2021screwnet,
  title={Screwnet: Category-independent articulation model estimation from depth images using screw theory},
  author={Jain, Ajinkya and Lioutikov, Rudolf and Chuck, Caleb and Niekum, Scott},
  booktitle={2021 IEEE International Conference on Robotics and Automation (ICRA)},
  pages={13670--13677},
  year={2021},
}

@inproceedings{Abbatematteo2019LearningTG,
  title={Learning to Generalize Kinematic Models to Novel Objects},
  author={Ben Abbatematteo and Stefanie Tellex and George Dimitri Konidaris},
  booktitle={Conference on Robot Learning},
  year={2019},
}

@inproceedings{EisnerZhang2022FLOW,

  title={FlowBot3D: Learning 3D Articulation Flow to Manipulate Articulated Objects},

  author={Eisner*, Ben and Zhang*, Harry and Held,David},

  booktitle={Robotics: Science and Systems (RSS)},

  year={2022}

}

@InProceedings{pmlr-v229-zhang23c,
  title = 	 {FlowBot++: Learning Generalized Articulated Objects Manipulation via Articulation Projection},
  author =       {Zhang, Harry and Eisner, Ben and Held, David},
  booktitle = 	 {Proceedings of The 7th Conference on Robot Learning},
  pages = 	 {1222--1241},
  year = 	 {2023},
}

@INPROCEEDINGS{4543220,
  author={Katz, Dov and Brock, Oliver},
  booktitle={2008 IEEE International Conference on Robotics and Automation}, 
  title={Manipulating articulated objects with interactive perception}, 
  year={2008},
  pages={272-277},}

@conference{Katz-2013-7694,
author = {Dov Katz and Moslem Kazemi and J. Andrew (Drew) Bagnell and Anthony (Tony) Stentz},
title = {Interactive Segmentation, Tracking, and Kinematic Modeling of Unknown 3D Articulated Objects},
booktitle = {Proceedings of (ICRA) International Conference on Robotics and Automation},
year = {2013},
pages = {5003 - 5010},
}

@article{nie2022sfa, 
            title={Structure from Action: Learning Interactions for Articulated Object 3D Structure Discovery}, 
            author={Nie, Neil and Gadre, Samir Yitzhak and Ehsani, Kiana and Song, Shuran},
            journal={arxiv preprint arXiv:2207.08997 },
            year={2022} }

@InProceedings{Xiang_2020_SAPIEN,
author = {Xiang, Fanbo and Qin, Yuzhe and Mo, Kaichun and Xia, Yikuan and Zhu, Hao and Liu, Fangchen and Liu, Minghua and Jiang, Hanxiao and Yuan, Yifu and Wang, He and Yi, Li and Chang, Angel X. and Guibas, Leonidas J. and Su, Hao},
title = {{SAPIEN}: A SimulAted Part-based Interactive ENvironment},
booktitle = {The IEEE Conference on Computer Vision and Pattern Recognition (CVPR)},
month = {June},
year = {2020}}

@InProceedings{Mo_2019_CVPR,
author = {Mo, Kaichun and Zhu, Shilin and Chang, Angel X. and Yi, Li and Tripathi, Subarna and Guibas, Leonidas J. and Su, Hao},
title = {{PartNet}: A Large-Scale Benchmark for Fine-Grained and Hierarchical Part-Level {3D} Object Understanding},
booktitle = {The IEEE Conference on Computer Vision and Pattern Recognition (CVPR)},
year = {2019}
}

@article{chang2015shapenet,
title={Shapenet: An information-rich 3d model repository},
author={Chang, Angel X and Funkhouser, Thomas and Guibas, Leonidas and Hanrahan, Pat and Huang, Qixing and Li, Zimo and Savarese, Silvio and Savva, Manolis and Song, Shuran and Su, Hao and others},
journal={arXiv preprint arXiv:1512.03012},
year={2015}
}

@article{vaswani2017attention,
  title={Attention is all you need},
  author={Vaswani, A},
  journal={Advances in Neural Information Processing Systems},
  year={2017}
}

@article{chen2024urdformer,
  title={URDFormer: A Pipeline for Constructing Articulated Simulation Environments from Real-World Images},
  author={Zoey Chen and Aaron Walsman and Marius Memmel and Kaichun Mo and Alex Fang and Karthikeya Vemuri and Alan Wu and Dieter Fox and Abhishek Gupta},
  journal={arXiv preprint arXiv:2405.11656},
  year={2024}
}

@inproceedings{jain2022distributional,
  title={Distributional depth-based estimation of object articulation models},
  author={Jain, Ajinkya and Giguere, Stephen and Lioutikov, Rudolf and Niekum, Scott},
  booktitle={Conference on Robot Learning},
  pages={1611--1621},
  year={2022},
}

@article{li2020categorylevelarticulatedobjectpose,
      title={Category-Level Articulated Object Pose Estimation}, 
      author={Xiaolong Li and He Wang and Li Yi and Leonidas Guibas and A. Lynn Abbott and Shuran Song},
      year={2020},
      journal = {arxiv preprint arXiv:1912.11913 },
}

@article{10.1145/3272127.3275027,
author = {Yi, Li and Huang, Haibin and Liu, Difan and Kalogerakis, Evangelos and Su, Hao and Guibas, Leonidas},
title = {Deep part induction from articulated object pairs},
year = {2018},
journal = {ACM Trans. Graph.},

}

@INPROCEEDINGS{7487714,
  author={Martín-Martín, Roberto and Höfer, Sebastian and Brock, Oliver},
  booktitle={2016 IEEE International Conference on Robotics and Automation (ICRA)}, 
  title={An integrated approach to visual perception of articulated objects}, 
  year={2016},
  pages={5091-5097},
  doi={10.1109/ICRA.2016.7487714}}

@article{geng2022gapartnet,
  title={GAPartNet: Cross-Category Domain-Generalizable Object Perception and Manipulation via Generalizable and Actionable Parts},
  author={Geng, Haoran and Xu, Helin and Zhao, Chengyang and Xu, Chao and Yi, Li and Huang, Siyuan and Wang, He},
  journal={arXiv preprint arXiv:2211.05272},
  year={2022}
}

@inproceedings{jiayi2023paris,
          author    = {Liu, Jiayi and Mahdavi-Amiri, Ali and Savva, Manolis},
          title     = {{PARIS}: Part-level Reconstruction and Motion Analysis for Articulated Objects},
          year      = {2023},
          booktitle = {Proceedings of the IEEE International Conference on Computer Vision (ICCV)}
      }

@article{qi2017pointnetplusplus,
      title={PointNet++: Deep Hierarchical Feature Learning on Point Sets in a Metric Space},
      author={Qi, Charles R and Yi, Li and Su, Hao and Guibas, Leonidas J},
      journal={arXiv preprint arXiv:1706.02413},
      year={2017}
    }

@article{kim2025screwsplat,
title={ScrewSplat: An End-to-End Method for Articulated Object Recognition},
author={Kim, Seungyeon and Ha, Junsu and Kim, Young Hun and Lee, Yonghyeon and Park, Frank C},
journal={arXiv preprint arXiv:2508.02146},
year={2025}
}

@inproceedings{mildenhall2020nerf,
 title={NeRF: Representing Scenes as Neural Radiance Fields for View Synthesis},
 author={Ben Mildenhall and Pratul P. Srinivasan and Matthew Tancik and Jonathan T. Barron and Ravi Ramamoorthi and Ren Ng},
 year={2020},
 booktitle={ECCV},
}

@Article{kerbl3Dgaussians,
      author       = {Kerbl, Bernhard and Kopanas, Georgios and Leimk{\"u}hler, Thomas and Drettakis, George},
      title        = {3D Gaussian Splatting for Real-Time Radiance Field Rendering},
      journal      = {ACM Transactions on Graphics},
      year         = {2023},
      
}

@inproceedings{weng2024neural,
  title={Neural implicit representation for building digital twins of unknown articulated objects},
  author={Weng, Yijia and Wen, Bowen and Tremblay, Jonathan and Blukis, Valts and Fox, Dieter and Guibas, Leonidas and Birchfield, Stan},
  booktitle={Proceedings of the IEEE/CVF Conference on Computer Vision and Pattern Recognition},
  pages={3141--3150},
  year={2024}
}

@article{yu2025artgs3dgaussiansplattinginteractive,
      title={ArtGS:3D Gaussian Splatting for Interactive Visual-Physical Modeling and Manipulation of Articulated Objects}, 
      author={Qiaojun Yu and Xibin Yuan and Yu jiang and Junting Chen and Dongzhe Zheng and Ce Hao and Yang You and Yixing Chen and Yao Mu and Liu Liu and Cewu Lu},
      year={2025},
      journal={arXiv preprint arXiv:2507.02600 },
}

@article{liu2025artgsbuildinginteractablereplicas,
      title={ArtGS: Building Interactable Replicas of Complex Articulated Objects via Gaussian Splatting}, 
      author={Yu Liu and Baoxiong Jia and Ruijie Lu and Junfeng Ni and Song-Chun Zhu and Siyuan Huang},
      year={2025},
      journal={arXiv preprint arXiv:2502.19459 },
}

@inproceedings{10.1007/978-3-030-58452-8_13,
author = {Carion, Nicolas and Massa, Francisco and Synnaeve, Gabriel and Usunier, Nicolas and Kirillov, Alexander and Zagoruyko, Sergey},
title = {End-to-End Object Detection with Transformers},
year = {2020},
booktitle = {Computer Vision – ECCV 2020: 16th European Conference, Glasgow, UK, August 23–28, 2020, Proceedings, Part I},
pages = {213–229},

}

@article{wang2021neus,
  title={Neus: Learning neural implicit surfaces by volume rendering for multi-view reconstruction},
  author={Wang, Peng and Liu, Lingjie and Liu, Yuan and Theobalt, Christian and Komura, Taku and Wang, Wenping},
  journal={arXiv preprint arXiv:2106.10689},
  year={2021}
}

@inproceedings{kerr2024rsrd,
 title={Robot See Robot Do: Imitating Articulated Object Manipulation with Monocular 4D Reconstruction},
 author={Justin Kerr and Chung Min Kim and Mingxuan Wu and Brent Yi and Qianqian Wang and Ken Goldberg and Angjoo Kanazawa},
 booktitle={8th Annual Conference on Robot Learning},
 year={2024},
}

@article{garrido2014automatic,
  title={Automatic generation and detection of highly reliable fiducial markers under occlusion},
  author={Garrido-Jurado, Sergio and Mu{\~n}oz-Salinas, Rafael and Madrid-Cuevas, Francisco Jos{\'e} and Mar{\'\i}n-Jim{\'e}nez, Manuel Jes{\'u}s},
  journal={Pattern Recognition},
  pages={2280--2292},
  year={2014},
  publisher={Elsevier}
}


\end{document}